# Detection of Anomalies in a Time Series Data using InfluxDB and Python

Anih John. T, Bede Chika Amadi, and Festus Chima Umeokpala

Institute für Data Science, Engineering, and Analytics.

TH Köln – University of Applied Sciences

February 2020

**Abstract:** Analysis of water and environmental data is an important aspect of many intelligent water and environmental system applications where inference from such analysis plays a significant role in decision making. Quite often these data that are collected through sensible sensors can be anomalous due to different reasons such as systems breakdown, malfunctioning of sensor detectors, and more. Regardless of their root causes, such data severely affect the results of the subsequent analysis. This paper demonstrates data cleaning and preparation for time-series data and further proposes cost-sensitive machine learning algorithms as a solution to detect anomalous data points in a time-series data. The following models: Logistic Regression, Random Forest, Support Vector Machines have been modified to support the cost-sensitive learning which penalizes misclassified samples thereby minimizing the total misclassification cost. Our results showed that Random Forest outperformed the rest of the models at predicting the positive class (i.e anomalies). Applying predictive model improvement techniques like data oversampling seem to provide little or no improvement to the Random Forest model. Interestingly, with recursive feature elimination we achieved a better model performance thereby reducing the dimensions in the data. Finally, with Influxdb and Kapacitor the data was ingested and streamed to generate new data points to further evaluate the model performance on unseen data, this will allow for early recognition of undesirable changes in the drinking water quality and will enable the water supply companies to rectify on a timely basis whatever undesirable changes abound.

**Keywords:** Imbalanced Classification, Time-series, Anomaly detection, Cost-sensitive Learning, Over-sampling, Machine learning

## 1. Introduction

With the level of technological advancement in the world today, increasingly, wireless sensors and actuators have become predominant in different fields of life. They are largely used in numerous monitoring systems and devices applied in manufacturing and machinery, airplanes and aerospace, cars, health care, robotics, environmental monitoring, and many other aspects of our day-to-day activities. Most applications that make use of sensors rely strongly on their accuracy. However, this is difficult to ascertain [1].

In addition to this, many sensors are also deployed under potentially harsh weather condition, making their breakage even more likely [1]. When a sensor becomes faulty in any form, it starts generating erroneous readings which could eventually lead to data corruption during transmission, hence, it is crucial to incorporate a means of checking the correctness of the generated data so as to smoothen out any irregularity therein. These irregularities could be referred to as an anomaly.

Anomaly detection can simply be referred to as the task of identifying the observations that do not conform to an expected pattern within a dataset [2]. Typically, these anomalous observations possess the ability of getting translated into problems such as structural defects, errors, or frauds, depending on the field or context from which they are being analysed. The deployment of effective machine learning models on a system for the purpose of anomaly detection plays a crucial role in enhancing the quality of such systems, hence saving it from further damage due to the abnormality.

This research paper presents the application of machine learning on a time series data. Since time series data is a data that changes over time, it is important we treat this differently from the conventional

Results might vary due to the stochastic nature of the ML learning algorithm. For source code: http://bit.ly/Anomaly-detection

machine learning problems. Machine learning is all about finding patterns in data; since time series data change over time, it turns out to be a useful pattern to utilize. Time series data consists of at least two things: first, an array of numbers that represents the data itself, and second, another array that contains a timestamp for each datapoints.

This paper is structured as follows:

- **Materials and methods:** This section presents the business description, the problem we are trying to solve using machine learning, data preparation methods for time-series data and proposed solution approach to the problem.
- **Results:** This section presents performance evaluation of the models and further methods to improve the best model's performance.
- **Discussion and Conclusion:** This section summarizes the paper's limitations and proposes ideas for future work in this area.

## 2. Materials and methods

This section will introduce the business, and the problem as well as several techniques to prepare the data for time series analysis and machine learning.

### 2.1 Business Understanding and Problem

The earth's surface is made up water to a total of 71% approximate proportion and this is very crucial for all forms of life [3]. Owing to the essential nature of water for sustainable living, water requires preservation knowing it is highly sensible to all kinds of contaminations. The provision of clean and safe drinking water is an essential task for water supply companies around the world [3].

To handle this situation, highly effective sensors monitor relevant water bodies and environmental data at several measuring points, daily. The monitored data can be analysed with aims to identify any kind of anomalies. This will allow for early recognition of undesirable changes in the drinking water quality and will enable the water supply companies to rectify on a timely basis whatever undesirable changes that abound [3].

### 2.2 Data Understanding

The data that was used for this research has been measured at different stations near the outflow of a designated waterworks. The data contains time series denoting water quality data. Table 1 gives an overview of the data provided [3]. The data contains 122,334 observations and 10 features, where *EVENT* would be used as the target variable. Inspecting the target variable, we see that it consists of approximately 98.5% as normal data points or majority class and approximately 1.42% as anomalies or minority class, clearly, this is a case of an *Imbalance Classification problem*.

**Table 1.** Description of the time series data [3]

| Column name | Description |
|---|---|
| Time | Time of measurement, given in following format: yyyy-mm-dd HH:MM:SS |
| Tp | The temperature of the water, given in °C. |
| Cl | Amount of chlorine dioxide in the water, given in mg/L (MS1) |
| pH | PH value of the water |
| Redox | Redox potential, given in mV |
| Leit | Electric conductivity of the water, given in $\mu S/cm$ |
| Trueb | Turbidity of the water, given in NTU |
| Cl_2 | Amount of chlorine dioxide in the water, given in mg/L (MS2) |
| Fm | Flow rate at water line 1, given in $m^3/h$ |
| Fm_2 | Flow rate at water line 2, given in $m^3/h$ |
| EVENT | Marker if this entry should be considered as a remarkable change resp. event, given in boolean. |

Results might vary due to the stochastic nature of the ML learning algorithm. For source code: http://bit.ly/Anomaly-detection

## 2.3 Data Cleaning

Data can be vast and thus always messy; it requires fine tuning and cleaning before fitting it to the models. In timeseries, messy data is often a product of failing sensors or human errors in logging the data, database failure and more.

The Figure 1 show the values of various features of the dataset from 15-02-2016 to 10-05-2016. There seems to be missing values between the periods 07-03-2016 and 17-04-2016, as well as some period where the data does not fluctuate at all.

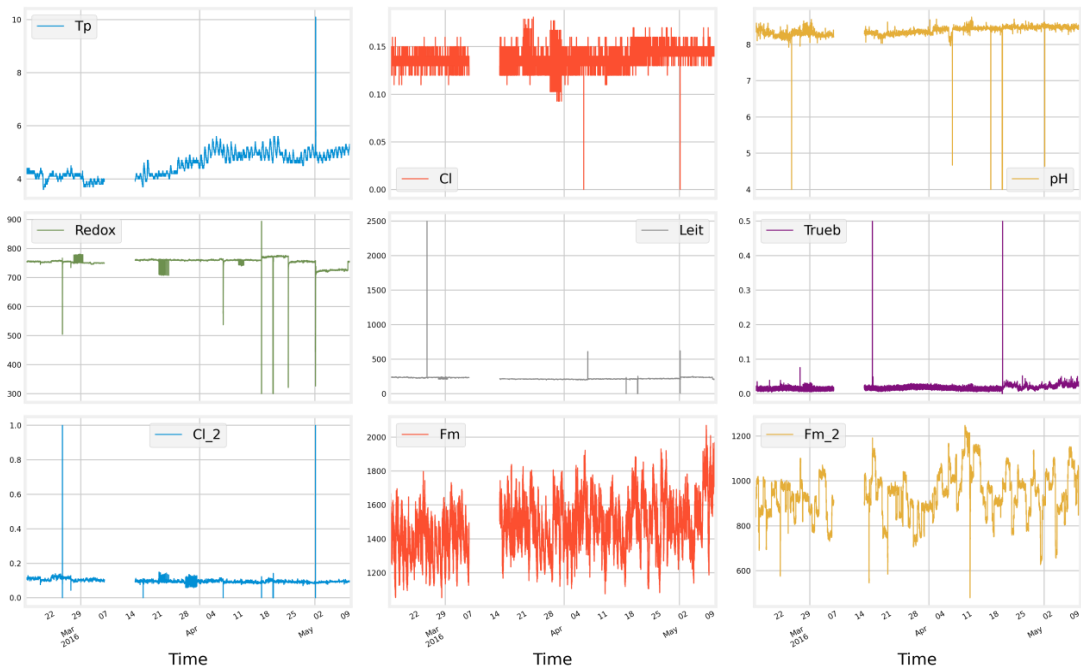

**Figure 1.** Data visualization of the time series data

Sadly, most predictive modeling techniques cannot handle any missing values. Thus, the problem of missing values must be addressed prior to data modeling. A common way to handle missing data is to interpolate the missing values. Interpolation refers to a technique of using known values on either end of a missing window of time to infer what is in between. While there are various techniques of interpolation, we have handled missing values using zero interpolation method, which infers the missing values by copying the last known value before the missing value in that column.

## 2.4 Data Preparation

For a time-series model, its trend is a very important feature. A positive trend is a line that generally slopes up; the values increases with time. Similarly, where the value decreases, it is termed a negative trend. Another important concept in a time series is white noise, this is a series of measurements where each value is uncorrelated with previous values. To effectively model a time series, it must be stationary-meaning that the distribution of the data does not change with time. When a time series is stationary, it becomes easier to model the data. In fact, statistical modelling methods assume that the time series is stationary for it to be very effective.

For a time-series to be stationary, it must fulfill three criteria:

1. The series has zero trend, it is not growing or shrinking,
2. The variance is constant.
3. The autocorrelation is constant.

The most common test for identifying whether a time series is stationary or not is the *augmented Dicky-Fuller test*. This is a statistical test where the null hypothesis assumes that the time series is non-stationary due to trend. Using the augmented Dicky Fuller test shows that the data is stationary but visualizing the data in Figure 2 proved otherwise, taking a difference of the data using equation (1) shows a more stationary plot of the data points in Figure 3, where the data is roughly centered at zero and the periods of high and low changes are easier to spot. Table 2 shows the results of the augmented Dicky-Fuller test after taking the difference. We explain the result using the p-value. A p-value below a threshold less than 1% implies that we reject the null hypothesis i.e. the time series is stationary. Having more negative values on the ADF statistics gives us more reason to reject the null hypothesis. Table 2 shows that the test statistic for all the variables is far less than the values at 1%. This clearly suggests that we can safely reject the null hypothesis with a significance level of less than 1% because the probability of getting a p-value as low as that by random chance is very unlikely.

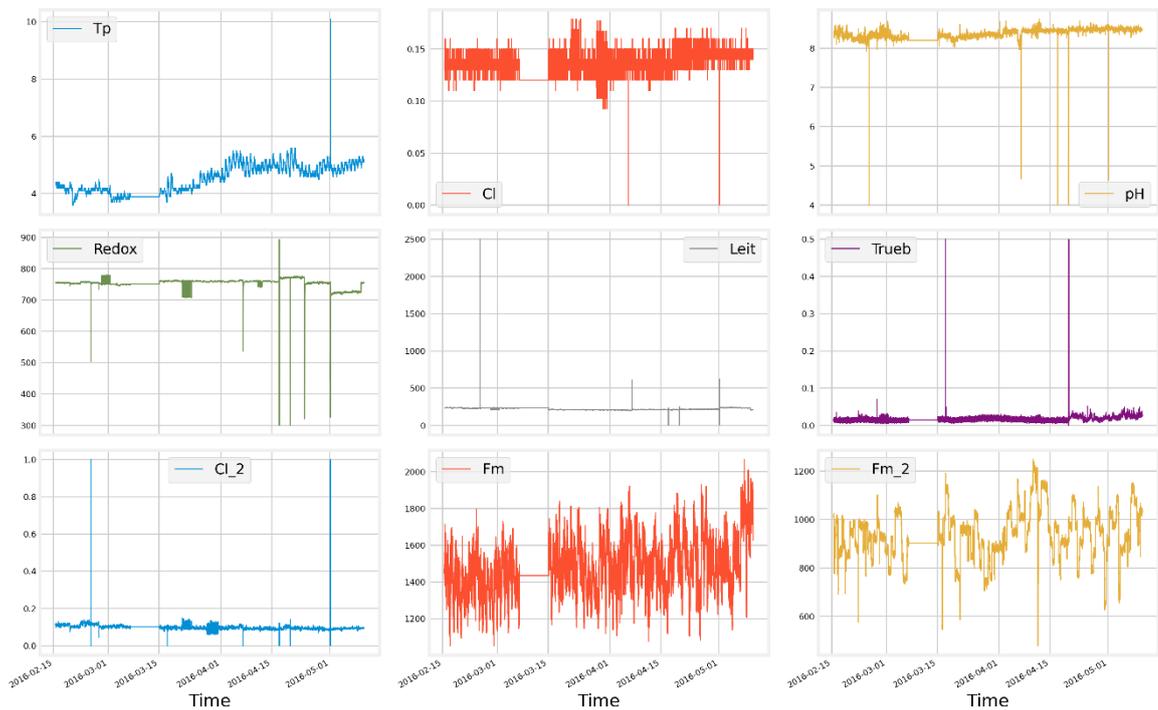

**Figure 2.** Data visualization of the data before applying differencing

**Difference:** $\nabla y_t = y_t - y_{t-1}$      **(1)**

**Table 2.** Results from augmented Dicky-Fuller test after differencing

|  | Tp | Cl | pH | Redox | Leit | Trueb | Cl_2 | Fm | Fm_2 |
|---|---|---|---|---|---|---|---|---|---|
| **ADF Statistic** | -65.6594 | -64.3677 | -65.1679 | -65.7522 | -64.892 | -65.0892 | -64.949 | -65.9133 | -64.8712 |
| **p-value** | 0.0000 | 0.0000 | 0.0000 | 0.0000 | 0.0000 | 0.0000 | 0.0000 | 0.0000 | 0.0000 |
| **1%** | -3.43042 | -3.43042 | -3.43042 | -3.43042 | -3.43042 | -3.43042 | -3.43042 | -3.43042 | -3.43042 |
| **5%** | -2.86157 | -2.86157 | -2.86157 | -2.86157 | -2.86157 | -2.86157 | -2.86157 | -2.86157 | -2.86157 |
| **10%** | -2.56679 | -2.56679 | -2.56679 | -2.56679 | -2.56679 | -2.56679 | -2.56679 | -2.56679 | -2.56679 |

Results might vary due to the stochastic nature of the ML learning algorithm. For source code: http://bit.ly/Anomaly-detection

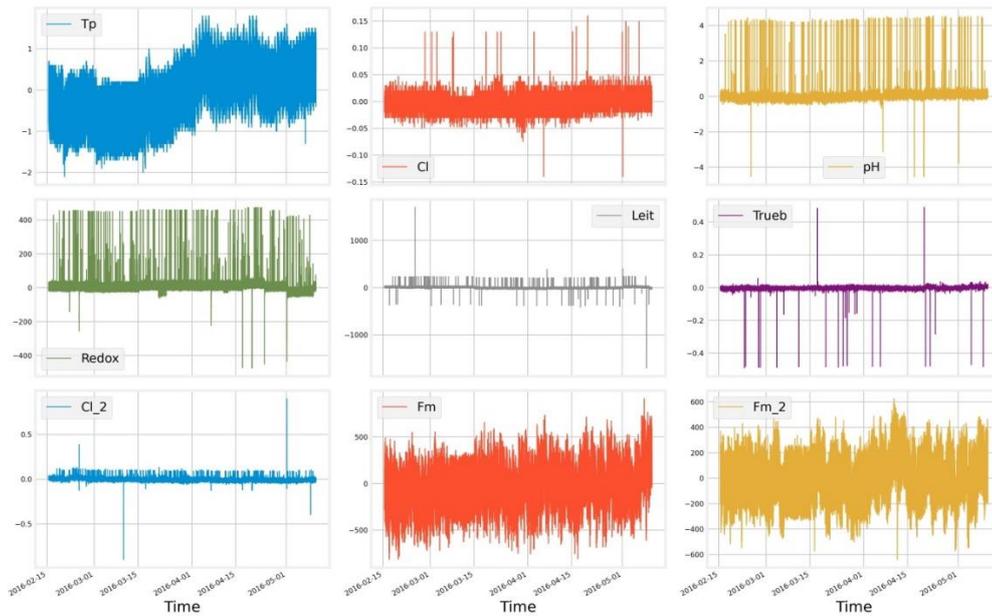

**Figure 3.** Data visualization of the stationary time series data

## 2.5 Feature selection and importance

Feature selection can be referred to as the process of selecting and identifying input features that are important to the target variable, by so doing we can reduce dimensionality of the features by discarding irrelevant features from the model and also maintain the model performance. While there are several feature selection methods, we have used mutual information statistics to get a feeling of our features before modelling, we have also explored more feature selection methods at later sections of this paper.

Mutual Information applies the concept of entropy to specify how much common certainty are present between two data features and measures the reduction in uncertainty for one feature given a known value of the other feature [5].

We perform feature selection with mutual information after we have prepared the data and ensured our data is stationary. The Figure 4 shows the feature importance using mutual information statistics. We see the most important feature in predicting our target is "*Redox*" and the least important feature is "*Tp*". Since there are different techniques for scoring features based on scores, which one do we rely on? In section 3, we have evaluated different models using different feature selection techniques and chosen the method that results in a model with the best performance.

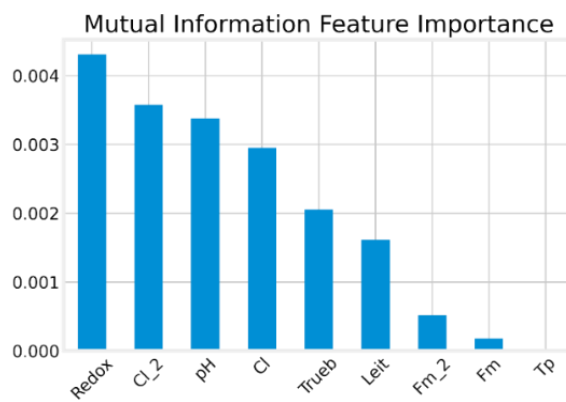

**Figure 4.** Bar Chart of the Input Features using Mutual Information Feature Importance

### 2.6 Proposed Machine Learning Algorithms

The field of machine learning that is focused on learning and using models with unequal costs when making prediction is referred to as Cost Sensitive Learning. Cost-Sensitive Learning makes use of cost matrix to vary the misclassification cost. Table 3 shows the cost matrix for cost sensitive learning.

Table 3. Cost Matrix for Binary Classification task

|  | Actual Negative | Actual Positive |
|---|---|---|
| **Predicted Negative** | C(0,0) True Negative | C(0,1) False Negative |
| **Predicted Positive** | C(1,0) False Positive | C(1,1) True positive |

Equation (2) below defines the total cost of a classifier as the cost-weighted sum of the False Negative and False Positives [4].

$$\text{Total Cost} = C(0,1) \times \text{FalseNegatives} + C(1,0) \times \text{FalsePositives} \quad (2)$$

Machine learning algorithms are not often developed specifically for cost-sensitive learning. Instead, the wealth of existing machine learning algorithms can be modified to make use of the cost matrix in Table 3 [4]. For this research work, the following machine learning algorithms have been used for the cost-sensitive learning: Logistic Regression, Support Vector Machines, and Random Forest.

### 2.7 Evaluation Metrics

Classification accuracy is defined as the number of accurate predictions divided by the total number of data points. While this evaluation metric is a commonly used metric, it will be an inappropriate evaluation metric for the problem we are trying to solve because the distribution of the target variable is not equal, and this might be misleading. The following metrics have been used for evaluation: Sensitivity-specificity metrics, F0.5 score and F1 Score. Sensitivity refers to the true positive rate and summarizes how well the positive class or the minority class were predicted [4]. Specificity is the complement to sensitivity, or the true negative rate, and summarizes how well the negative class or the majority class were predicted [4]. Since we are dealing with an imbalanced dataset, the sensitivity would be more important than the specificity. The F-measure metric is commonly used where a balance of precision and recall is required. Precision is the number of true positives divided by the total number of true positives and false positives. In our case, it is number of correctly labeled anomalies divided by the total number of data points classified as anomaly. Recall on the other hand is also called sensitivity which we have explained at the beginning of this section. Unlike the F-measure, the Fbeta-measure; an abstraction of the F-measure focuses on giving more attention to either precision or recall. It is controlled by a coefficient *beta*, in the calculation of the harmonic mean of precision-recall balance[4]. If the false positives are more costly, the F0.5 score(*beta = 0.5*) would be preferred. If both false positives and false negatives are equally costly, the F1 score(*beta = 1*) would be preferred.

**Sensitivity-specificity Metrics**

$$\text{Sensitivity} = \frac{TruePositive}{TruePositive+FalseNegative} \quad (3) \quad \text{Specificity} = \frac{TrueNegative}{FalsePositive+TrueNegative} \quad (4)$$

**F-measure Metrics**

$$\text{F-measure} = \frac{2 \; x \; Precision \; x \; Recall}{Precision+Recall} \quad (5) \quad \text{Fbeta-measure} = \frac{(1+\beta^2) \; x \; Precision \; x \; Recall}{\beta^2 \; x \; Precision+Recall} \quad (6)$$

Results might vary due to the stochastic nature of the ML learning algorithm. For source code: http://bit.ly/Anomaly-detection

## 3. Results

### 3.1 Data Modelling and Evaluation

### 3.1.1 Evaluate Cost-Sensitive Algorithms

Logistic Regression, Support Vector Machines and Random Forest which support Cost-Sensitive Learning have been evaluated individually using *Repeated Stratified k-fold Cross Validation*. This procedure provides a good general estimate of model performance that is not too optimistically biased compared to a single train-test split [4]. We have used k=10, meaning each fold will contain about $\frac{122334}{10}$ or about 12,233 samples. Stratified implies that each fold will contain the same mixture of examples by class, that is approximately 99 percent to 1 percent of normal to anomaly examples. Repetition indicates that the evaluation process will be performed multiple times to help avoid fluke results and better capture the variance of the chosen model. In this, we have specified three repeats. That means a single model will be fitted and evaluated 10 x 3 (30) times, the mean and standard deviation of the errors are used. The results are reported in Figure 5 and Table 4. We see that the Random Forest model with 1000 Decision Trees outperformed the rest of the models with an F1 Score of approximately 86% and an F0.5 score of approximately 92%. In the next section we explored some techniques to improve our model performance.

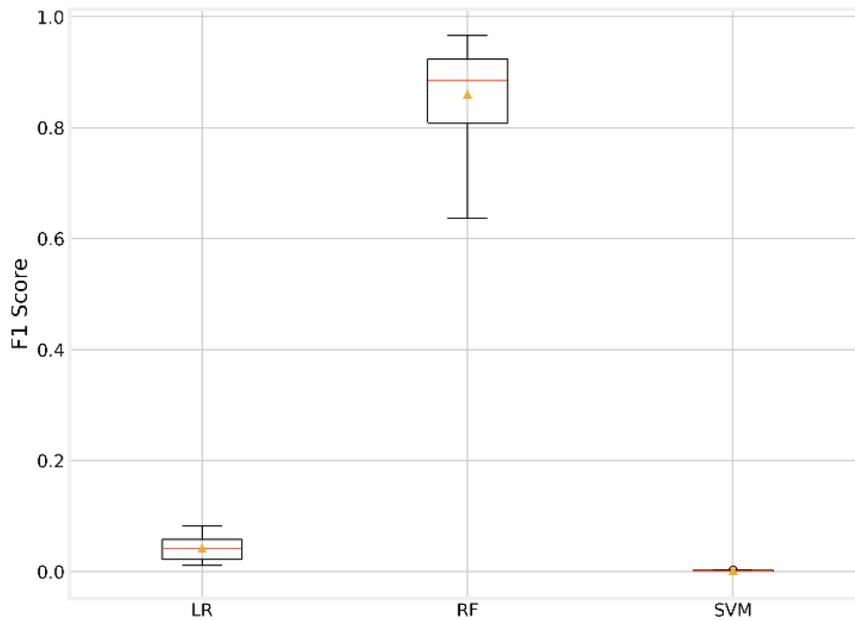

**Figure 5.** Box and Whisker Plots of Cost-Sensitive Machine Learning Models on the Imbalanced data

**Table 4.** Results from the Cost-Sensitive Machine Learning Models

| Models | Sensitivity | Specificity | F1-Score | F0.5 Score | Parameters |
|---|---|---|---|---|---|
| **Random Forest (RF)** | 78.6% | 100% | 86% | 92% | 1000 Decision Trees, balanced class weight, and the rest are default. |
| **Logistic Regression (LR)** | 49% | 97% | 4% | 2% | Balanced class weight and the rest are default |
| **Support Vector Machine (SVM)** | 31% | 56% | 0.2% | 0.1% | Balanced class weight, 1000 number of iterations and the rest are default |

Results might vary due to the stochastic nature of the ML learning algorithm. For source code: http://bit.ly/Anomaly-detection

### 3.1.2 Evaluate Data Oversampling Algorithms

Data sampling is another approach to address imbalanced data. It is also a good way to prepare the data before training a model. While there are various techniques to balance the class distribution for an imbalanced dataset, we focused on the oversampling algorithms as an approach to create duplicates from the minority class to balance the data. The following oversampling methods have been used with Random Forest to further improve the model's performance: Random Oversampling (ROS), Synthetic Minority Oversampling Technique (SMOTE), BorderLine SMOTE (BLSMOTE), SVM SMOTE, and Adaptive Synthetic Sampling (ADASYN).

Random Oversampling (ROS) creates duplicates from the minority class in order to balance the data, it is called random because it performed randomly. Synthetic Minority Oversampling Technique (SMOTE) selects samples that are close in the feature space by drawing a line between the samples in the feature space and drawing a new sample at a point along that line. [4]. Borderline-SMOTE is an extension to SMOTE, and it involves selecting those instances of the minority class that are misclassified with a K-Nearest Neighbor (KNN) classification model and subsequently oversample only those difficult cases. Support Vector Machine (SVM) SMOTE is an alternative approach to Borderline-SMOTE, in this case, an SVM is used instead of a K-Nearest neighbor model, and finally the Adaptive Synthetic Sampling (ADASYN) creates new samples inversely proportional to the density of the examples in the minority class i.e. It creates more synthetic examples in the regions of the feature space where the density of minority examples are relatively low, and little or no examples where the densities are relatively high [4].

We have performed an experiment with these algorithms with their default configuration. Figure 6 shows the results of the F1-scores. We see a little improvement of the F1 score with Random sampling of the data from 86% to 87.6% and an F0.5 score of 91.5% which is approximately the same with the baseline. Now let us see how we could improve the model further with other methods.

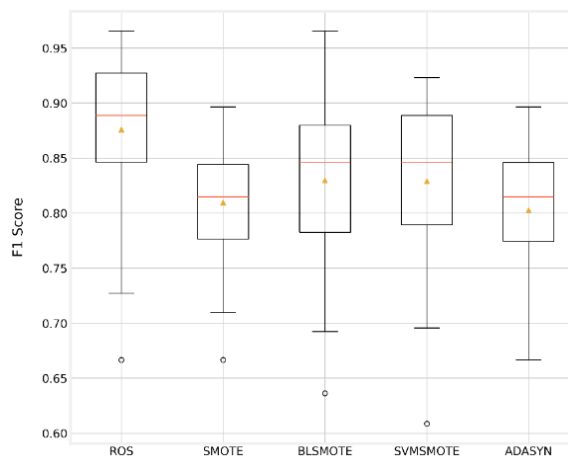

**Figure 6.** Box and Whisker Plot of Random Forest models With Data Oversampling on the Imbalanced Data.

### 3.1.3 Recursive Feature Elimination

Recursive Feature Elimination (RFE) is a popular feature selection algorithm, as the name implies, it removes unimportant features from the model recursively [5]. In section 2.4, we discussed briefly about mutual information statistics and the problem of selecting appropriate scoring features. Using the best performing model, with RFE to get rid of features that do not increase the performance of the model, by so doing we increase the model's F1 score and reduce the dimensions in our data. Figure 7 (a) shows the boxplot with the mean samples of the F1 Scores with the number of features selected on the x-axis. 7 (b) shows the rankings of each feature. We see that the feature *"Tp"* has the lowest ranking, interestingly after the feature "Tp" was added, the F1-score reduced slightly, this is also evident in the Figure 7 (a). We can safely conclude that the feature *"Tp"* does not constitute much information to our target variable, therefore we exclude *"TP"* from our input feature to reduce the noise in our model and improve the model performance.

Results might vary due to the stochastic nature of the ML learning algorithm. For source code: http://bit.ly/Anomaly-detection

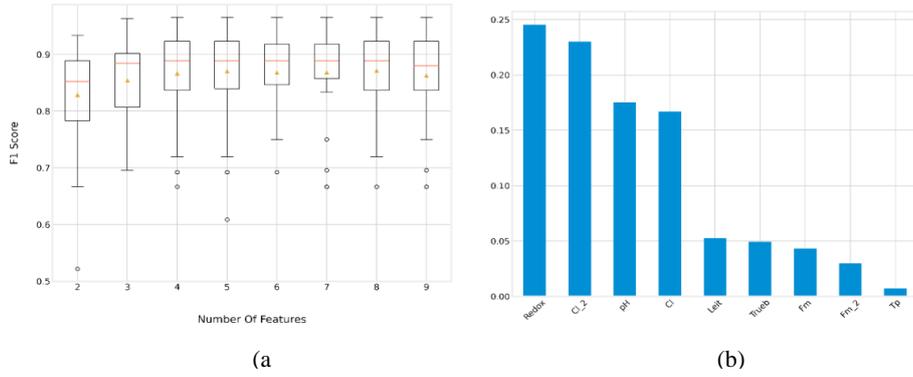

**Figure 7.** BoxPlot of RFE Number of Selected Features vs. F1 Score.

### 3.2 Data Ingestion and Streaming with InfluxDB and Kapacitor

InfluxDB is an open source time series database that is part of the TICK (Telegraf, InfluxDB, Chronograf, Kapacitor) stack [6]. Like every other database system, it was designed to write high quality queries to query data and store data. Kapicitor which is a part of TICK, was designed to stream data points in real time. Kapacitor utilizes a Domain Specific Language (DSL) named TICKscript to define tasks involving the extraction, transformation and loading of data (ETL), likewise, the tracking of arbitrary changes and the recognition of events within data [7]. The steps followed to achieve this are discussed as follows:

First, we defined the alerts. TICKscript is used in .tick files to define ETL pipelines for processing data [7]. The TICKscript language is intended to chain together the invocation of data processing operations defined in nodes [7]. In the TICK script, we specified a period of 5 days and 2 hours for the generation of new data points i.e. every 2 hours, 7200 data points are generated. Additionally, the TICK script contains two edges which are:

- stream→from()– which defines the processing type of the task and the data stream [7].
- from()→httpOut()– which passes the data stream to the HTTP output processing node [7].

Secondly, with python, we write the raw data to influxdb and enable the TICK script with a type stream to stream the data in real time to generate new data points.

Finally, with the httpout specified in our TICK Script, we retrieve the new data points. The new data are streamed to the HTTPOUT specified in the TICK script and it is available in a JSON structure. With python, we read the data from the HTTPOUT and load the data with pandas json method, and also load our working model and make predictions on the new data points to capture outliers in the data points.

The Figure 8 shows the overall picture of the process described.

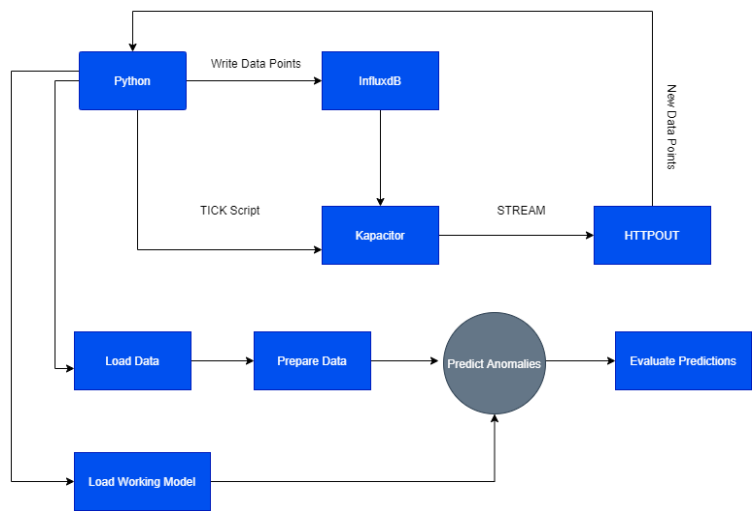

**Figure 8.** Flow diagram of the data and working process

Results might vary due to the stochastic nature of the ML learning algorithm. For source code: http://bit.ly/Anomaly-detection

## 4. Discussions and conclusion

This paper proposed cost-sensitive machine models as a solution to an imbalanced time-series classification task to detect anomalies from data generated through sensors.

First, we started by cleaning the data and preparing the data for machine learning, since we are dealing with a time-series data, we ensured that the data was stationary to avoid long-term drift. Part of the goal of every machine learning model is that the built model avoids overfitting, that is, the model is able to generalize well to an unseen data. As a customary step the data was split, with some data held as validation data.

Furthermore, we evaluated and compared the performance of our cost sensitive machine learning models using a boxplot that shows the mean of the repeated stratified k-fold cross validation approach, this was done to ensure a good estimate for our models. Our experiments demonstrated that Random Forest with 1000 Decision Trees outperformed the rest of the Cost-Sensitive Machine Learning algorithms. The prediction from the best performing model - Random Forest, for the most important feature *"Redox"* can be visualized in Figure 9, with the detected outliers in red and data points for the Redox variable in blue.

We investigated other strategies like data oversampling methods to see if we could further improve the model performance. However, the methods did not produce any significant increase in the performance of our models and it was computationally expensive to run them.

Additionally, we explored feature selection algorithms to investigate what features actually contributes towards the prediction of our target variable, it was noted that the variable *"Tp"* did not provide any information towards our target variable, hence we excluded the variable from our input feature by so doing we increased the model performance.

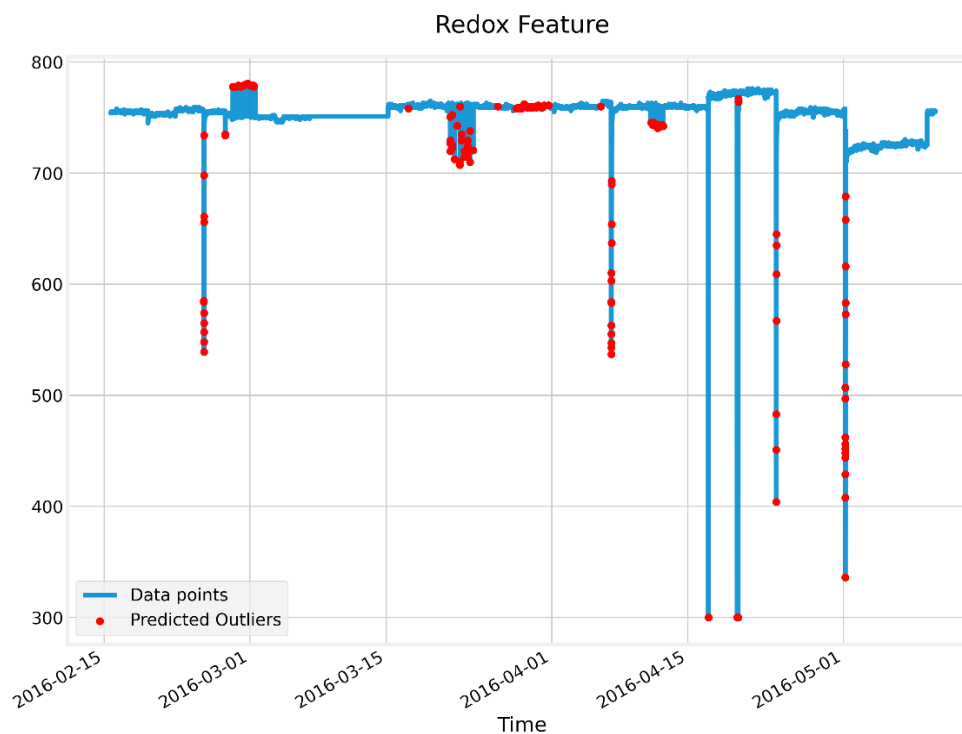

**Figure 9.** Random Forest predictions for the "Redox" column

Results might vary due to the stochastic nature of the ML learning algorithm. For source code: http://bit.ly/Anomaly-detection

The following evaluation metrics have been used for the model evaluation: Sensitivity, Specificity, F1 score and F0.5 score. We came to the following conclusions:

- If the false positives are more costly, the F0.5 score should be used for evaluation.

- If both false positives and False negatives are equally costly, the F1 score should be used for the evaluation.

- If predicting the normal data points is very important, the specificity score would be an interesting metric to look out for, similarly if the anomalous points are very costly, the sensitivity score would be an interesting metric to watch out for.

We concluded this paper by performing data ingestion and streaming with influxdb and Kapacitor via a TICK script to specify the stream interval of 5 days at an instance of every 2 hours and then load the streamed data points from the HTTPOUT with python and load the working model to predict outliers. Overall, our model proved effectively at predicting outliers in the sensor data.

## 5. Future Works

This paper has captured the detection of anomalies in a time series data using cost-sensitive machine learning algorithms by approaching the problem like a supervised machine learning task. While we seem to have an effective model that does well at predicting outliers in the sensors data, there are still a lot more that could be done to further improve the model performance.

Hyperparameters are parameters we set before the training the model. Performing hyperparameter tuning could further improve model performance. For this project, we built a Random Forest of 1000 trees. Since Random Forest is a combination of decision trees to make decisions, what if we could increase the number of trees. There are also several hyperparameter tuning algorithms that could be applied to improve the model like Bayesian optimization, Grid Search, Random Search, Combination of Random Search and Grid Search, since this is computationally expensive, this was one of our limitations to this project.

This paper treated this task like a supervised machine learning task; however, this problem could be treated as an unsupervised machine learning task and outlier-based machine learning algorithms could be explored. Some of the outlier-based machine learning algorithms are Isolation Forest, Local Outlier Factor, One-Class SVM (Support Vector Machines) and more [8].

Since data was ingested and streamed via Influxdb and Kapcitor, another thing that could be improved is monitoring the data to avoid data drift at a later time.

Finally, Influxdb provides a platform for outlier detection. Interfacing this technology with the proposed machine learning algorithm and having the detection automated and happening in real time would be an interesting project that could be looked into.

## Acknowledgements

Special thanks to Dr. Martin Zaefferer, who supervised this project for his insights and guidance.

## References

1. F. Giannoni, M. Mancini, and F. Marinelli., "Anomaly detection models for IoT time series data," *Data Mining - Project Report* 2018, Accessed: Jan.11, 2020 [Online] Available: https://www.researchgate.net/-publication/329387705
2. Naveen Joshi, "Machine learning for anomaly detection," Accessed: Jan.10, 2020. [Online]. Available: https://www.allerin.com/blog/machine-learning-for-anomaly-detection

Results might vary due to the stochastic nature of the ML learning algorithm. For source code: http://bit.ly/Anomaly-detection

3. Sowmya Chandrasekaran, Martina Friese, Jörg Stork, Margarita Rebolledo, Thomas Bartz-Beielstein, "GECCO 2017 Industrial Challenge Monitoring of drinking-water quality," Accessed: Dec.03, 2020. [Online]. Available: http://www.spotseven.de/gecco/gecco-challenge/gecco-challenge-2017/
4. Jason Brownlee, "Imbalanced Classification with Python: Choose Better Metrics, Balance Skewed Classes, and Apply Cost-Sensitive Learning," Vermount Victoria, Australia. Machine Learning Mastery Pty Ltd,2020, pp.36-46,178-220.
5. Jason Brownlee, Data Preparation for Machine Learning: Data Cleaning, Feature Selection, and Data Transforms in Python," Vermount Victoria, Australia. Machine learning Mastery Pty Ltd,2020, pp.119, pp.150-154
6. "InfluxDB", InfluxData Inc, San Francisco, CA, USA. Accessed: Feb.21,2020. [Online]. Available: https://www.influxdata.com/products/influxdb-overview/
7. "Influxdata documentation," InfluxData Inc, San Francisco, CA, USA. Accessed: Feb.21,2020. [Online] https://docs.influxdata.com/kapacitor/v1.5/tick/introduction/
8. Jason Brownlee, "4 Automatic Outlier Detection Algorithms in Python," Accessed: Feb.10, 2020. [Online]. Available: https://machinelearningmastery.com/model-based-outlier-detection-and-removal-in-python/

Results might vary due to the stochastic nature of the ML learning algorithm. For source code: http://bit.ly/Anomaly-detection